\documentclass{article} 
\usepackage{iclr2015,times}
\usepackage{hyperref}
\usepackage{url}
\usepackage{float}
\usepackage{layout}
\usepackage{amsmath}
\usepackage{amsfonts}
\usepackage{graphicx}
\usepackage{multirow}
\usepackage{makecell}
\usepackage[export]{adjustbox}

\newcommand{\comment}[1]{}
\newcommand{\bx}{{\mathbf{x}}}
\newcommand{\btheta}{\mbox{\boldmath $\theta$}}

\title{Contour Detection Using Cost-Sensitive Convolutional Neural Networks}

\author{
Jyh-Jing Hwang \& Tyng-Luh Liu \\
Institute of Information Science\\
Academia Sinica\\
Taipei, Taiwan \\
\texttt{\{jyhjinghwang,liutyng\}@iis.sinica.edu.tw}
}

\iclrfinalcopy 

\begin{document}

\maketitle

\begin{abstract}
We address the problem of contour detection via per-pixel classifications of edge point. To facilitate the process, the proposed approach leverages with DenseNet, an efficient implementation of multiscale convolutional neural networks (CNNs), to extract an informative feature vector for each pixel and uses an SVM classifier to accomplish contour detection. The main challenge lies in adapting a pre-trained per-image CNN model for yielding per-pixel image features. We propose to base on the DenseNet architecture to achieve pixelwise fine-tuning and then consider a cost-sensitive strategy to further improve the learning with a small dataset of edge and non-edge image patches. In the experiment of contour detection, we look into the effectiveness of combining per-pixel features from different CNN layers and obtain comparable performances to the state-of-the-art on BSDS500.
\end{abstract}

%
\section{Introduction}
\label{sec:intro}
%

Contour detection is fundamental to a wide range of computer vision applications, including image segmentation~\citep{malik2001contour,gpb}, object detection~\citep{zitnick2014edge} and recognition~\citep{shotton2008multiscale}. The task is often carried out by exploring local image cues, such as intensity, color gradients, texture or local structures~\citep{canny,martin2004learning,mairal2008discriminative,gpb,se}. Take, for example, that ~\citet{se} use structured random forests to learn local edge patterns, and report current state-of-the-art results with impressive computation efficiency. More recently, object cues are also considered in ~\citet{deepnet} and ~\citet{n4} to further boost the performance. Despite the constant evolvement of relevant techniques in better solving the problem, seeking an appropriate feature representation remains the cornerstone of such efforts. We are thus motivated to propose a new learning formulation that can generate suitable per-pixel features for more satisfactorily performing contour detection.

We consider deep neural networks to construct a desired per-pixel feature learner. In particular, since the underlying task is essentially a classification problem, we adopt deep convolutional neural networks (CNNs) to establish a discriminative approach. However, one subtle deviation from typical applications of CNNs should be emphasized. In our method, we intend to use the CNN architecture, e.g., AlexNet~\citep{alexnet}, to generate features for each image pixel, not just a single feature vector for the whole input image. Such a distinction would call for a different perspective of parameter fine-tuning so that a pre-trained per-image CNN on ImageNet \citep{Deng2009imagenet} can be adapted into a new model for per-pixel edge classifications. To further investigate the property of the features from different convolutional layers and from various ensembles, we carry out a number of experiments to evaluate their effectiveness in performing contour detection on the benchmark BSDS Segmentation dataset~\citep{bsds}.

The organization of the paper is as follows. Section~\ref{sec:related} includes related work of contour detection and deep convolutional neural networks. In Section~~\ref{sec:model}, we describe the overall model for learning per-pixel features and useful techniques for fine-tuning the parameters. Section~\ref{sec:exp} provides detailed experimental results and comparisons to demonstrate the advantages of our method.  In Section~\ref{sec:discussion} we discuss the key ideas of the proposed techniques and possible future research efforts.

%
\section{Related Work}
\label{sec:related}
%

As stated, we focus on using a deep convolutional neural network to achieve feature learning for improving contour detection. The survey of relevant work is thus presented to give an insightful picture of the recent progress in each of the two areas of emphasis.

\subsection{Contour Detection}
\label{ssec:contour_detection}

Early techniques for contour detection~\citep{fram1975quantitative,canny,perona1990scale} mainly concern local image cues, such as intensity and color gradients. Amongst them, the Canny detector~\citep{canny} stands out for its simplicity and accuracy owing to exploring the peak gradient magnitude orthogonal to the contour direction. Detailed discussions about these approaches can be found in, e.g., \citet{bowyer1999edge}. Subsequent work along this line~\citep{martin2004learning,mairal2008discriminative,gpb} also identifies that textures are useful local cues for increasing the detection accuracy.

Apart from detecting local cues, learning-based techniques form a notable group in addressing this intriguing task~\citep{dollar2006supervised,mairal2008discriminative,zheng2010detecting,scg,sketchtokens,se,deepnet,n4}. \citet{dollar2006supervised} adopt a boosted classifier to independently label each pixel by learning its surrounding image patch. \citet{zheng2010detecting} analyze the combination of low-, mid-, and high-level information to detect object-specific contours. In addition, \citet{scg} propose to compute sparse code gradients and successfully improve \citet{gpb}.

\citet{sketchtokens} classify edge patches into sketch tokens using random forest classifiers, which can capture local edge structures. \citet{pmi} consider pointwise mutual information to extract global object contours. Their results display crisp and clean contours. Like in \citet{sketchtokens}, \citet{se} use structured random forests to learn edge patches, and achieve current state-of-the-art results in both accuracy and efficiency.

More relevant to our approach, \citet{deepnet} and \citet{n4} learn contour information with deep architectures. \citet{deepnet} encode and decode contours using multi-layer mean-and-covariance restricted Boltzmann machines. \citet{n4} establish a deep architecture, which composes of convolutional neural networks and nearest neighbor search, and obtain convincing results. Different from \citet{n4}, we strive for designing fine-tuning mechanisms with a small dataset for adapting an ImageNet pre-trained convolutional neural network for producing per-pixel image features. As we will see later, this effort leads to the state-of-the-art results of contour detection on the benchmark testing.

\subsection{Convolutional Neural Networks}
\label{ssec:CNNs}

Noticeably, CNNs are popularized by LeCun and colleagues who first apply CNNs to digit recognition~\citep{lecun1989backpropagation}, OCR~\citep{lecun1998gradient} and generic object recognition~\citep{jarrett2009best}. In contrast to using hand-crafted features, CNNs learn discriminative features and exhibit hierarchical semantic information along their deep architecture.

The AlexNet by \citet{alexnet} is perhaps the most popular implementation of CNNs for generic object classification. The model has been shown to outperform competing approaches based on traditional features in solving a number of mainstream computer vision problems. In \citet{turaga2010convolutional} and \cite{briggman2009maximin}, CNNs are used for image segmentation. To extend CNNs for object detection, \citet{farabet2013learning} utilize CNNs for semantic segmentation. \citet{sermanet2013overfeat} use CNNs to predict object locations via sliding window, while learning multi-stage features of CNNs for pedestrian detection is proposed in \citet{sermanet2013pedestrian}. \citet{girshick2013rich} also consider features from a deep CNN in a region proposal framework to achieve state-of-the-art object detection results on the PASCAL VOC dataset.

While CNNs thrives in generic object recognition and detection, less attention is paid to applications demanding per-pixel processing, such as contour detection and segmentation. Our method exploits the AlexNet model for contour detection and explores its per-pixel fine-tuning with a small dataset. Recently and independently from our work, generating per-pixel features based on CNNs can also be found in~\citet{hariharan2014hypercolumns} and \citet{long2014fully}.

\begin{figure}[th]
\begin{center}
\includegraphics[width=0.98\linewidth]{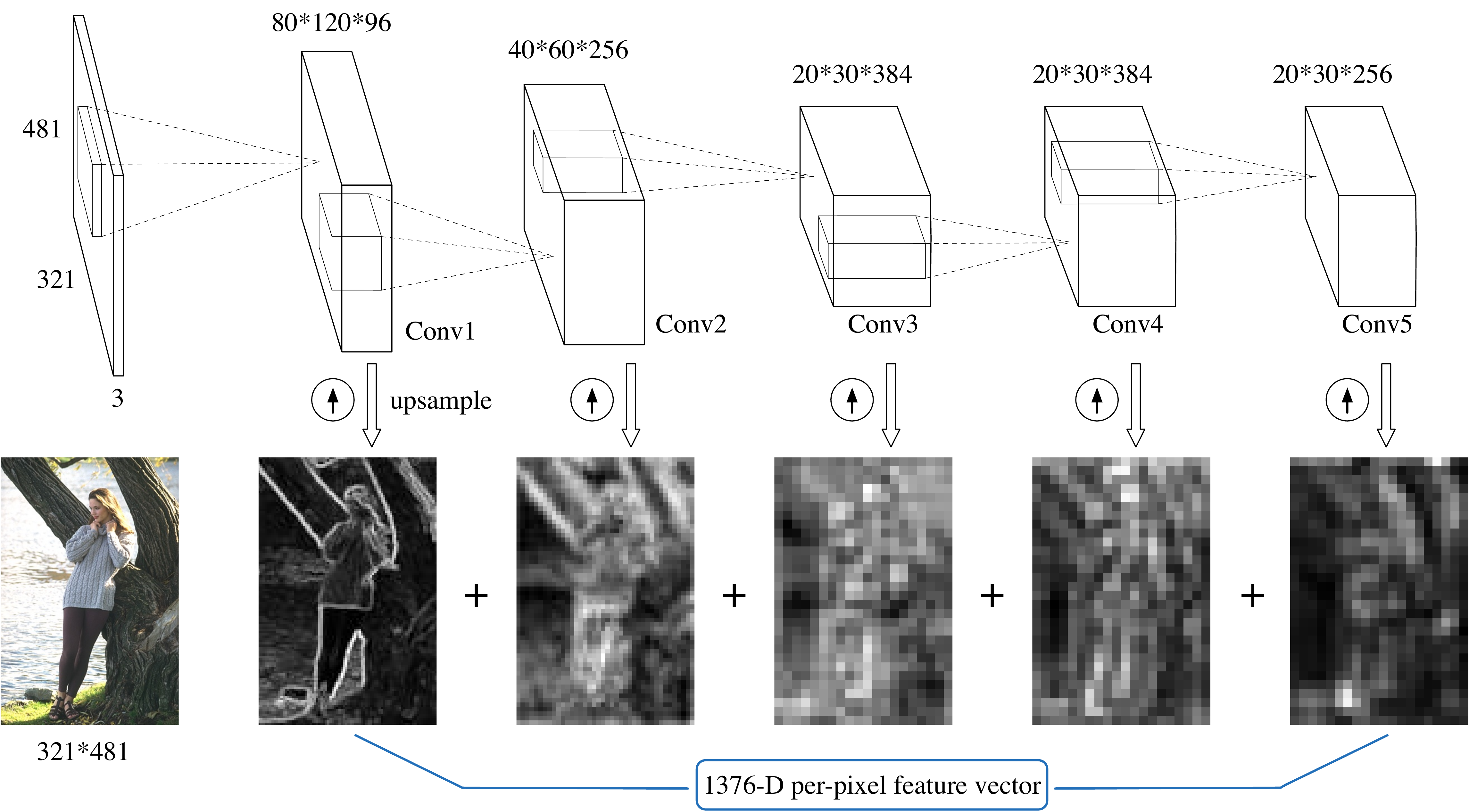}
\end{center}
\caption{Stacking feature maps from Conv1 to Conv5 yields a 1376-D feature vector per pixel.}
\label{fig:DCNN}
\end{figure}

%
\section{Per-Pixel CNN Features}
\label{sec:model}
%

Learning features by employing a deep architecture of neural net has been shown to be effective, but most of the existing techniques focus on yielding a feature vector for an input image (or image patch). Such a design may not be appropriate for vision applications that require investigating image characteristics in pixel level. In the problem of contour detection, the central task is to decide whether an underlying pixel is an edge point or not. Thus, it would be convenient that the deep network could yield per-pixel features.

We propose to construct a multiscale CNN model for contour detection. To this end, we extract per-pixel CNN features in AlexNet~\citep{alexnet} using DenseNet~\citep{densenet}, and pixelwise concatenate them to feed into a support vector machine (SVM) classifier. In particular, DenseNet provides fast multiscale feature pyramid extraction of any Caffe convolutional neural networks~\citep{caffe} and the convenience of working with images of arbitrary size. To extract per-pixel features, we upsample the feature maps from the first convolutional layer (Conv1) to the fifth convolutional layer (Conv5) to the original size of the input image. We then pixelwise stack the features from differen convolutional layers to constitute the per-pixel features. Depending on the selection of the convolutional layers, the resulting feature vector at each pixel would encode different level of information about an underlying pixel. Figure~\ref{fig:DCNN} illustrates the case of concatenating features from all five convolutional layers to form a 1376-D feature vector at each pixel.

To decide a pixel, say, $(i,j)$ is a contour point, one can now readily feed its corresponding feature vector to an SVM classifier. In practice, it is useful to include information from neighboring pixels so that local contour structures can be better distinguished. We consider the following eight neighboring pixels $(i\pm k,j), (i, j\pm k), (i\pm k, j \pm k)$ and append, starting from $(i-k, j-k)$, their respective feature vector to that of $(i,j)$ in clockwise order. In our implementation, we have tested $k=1, 2, 3$, which correspond to an image patch of size $3 \times 3$, $5 \times 5$ and $9 \times 9$, respectively.

\subsection{DenseNet Feature Pyramids}
\label{ssec:baseline}

We use DenseNet for CNN feature extraction because of its efficiency, flexibility, and availability. DenseNet is an open source system that computes dense and multiscale features from the convolutional layers of a Caffe CNN based object classifier. The process of feature extraction proceeds as follows. Given an input image, DenseNet computes its multiscale versions and stitches them to a large plane. After processing the whole plane by CNNs, DenseNet would unstitch the descriptor planes and then obtain multiresolution CNN descriptors.

The dimensions of convolutional features are ratios of the image size, e.g., one-fourth for Conv1, and one-eighth for Conv2. We rescale feature maps of all the convolutional layers to the image size. That is, there is a feature vector in every pixel. As illustrated in Figure~\ref{fig:DCNN}, the dimension of the resulting feature vector is $1376\times1$, which is concatenated by Conv1 ($96\times1$), Conv2 ($256\times1$), Conv3 ($384\times1$), Conv4 ($384\times1$), and Conv5 ($256\times1$).

For classification, we first concatenate features from the surrounding eight pixels to incorporate information about the local contour structure, and then use the combined per-pixel feature vectors to train a binary linear SVM. Specifically, in our multiscale setting, we train the SVM based on only the original resolution. In test time, we classify test images using both the original and the double resolutions. We average the two resulting edge maps for the final output of contour detection.

\subsection{Per-pixel Fine-tuning}
\label{ssec:per-pixel}

To fine-tune parameters for per-pixel contour detection, we exclude the two fully-connected layers of the ImageNet pre-trained CNN model in that the two layers will cause to restrict the input image size and consequently the overall architecture. We keep only the five convolutional layers, and on top of Conv5, we add a new 2-way softmax layer for edge classification.

Specifically, the input image size of ImageNet pre-trained CNN model is $227\times 227$, which is not suitable for our per-pixel design as each map in the Conv5 layer would still be $13\times 13$. In addition, we need to remove padding in CNN to conform to that DenseNet does not use padding (except the input plane). To carry out per-pixel fine-tuning, we first generate a set of edge and non-edge patches. The image (patch) size is set to $163\times 163$, and would reduce to $1 \times 1$ in Conv5, at which the 2-way softmax layer can now properly compute the per-pixel probability of being a contour point. Note that the loss for back-propagation is computed by the label prediction and the ground truth of the center pixel of $163\times 163$ input patch.

\subsection{Cost-sensitive Fine-tuning}
\label{ssec:cost-sensitive}

Compared with the number of parameters in DenseNet, the size of the training set of edge and non-edge patches is relatively small. Using the aforementioned per-pixel fine-tuning alone is usually insufficient to achieve good performance. Still, when addressing edges, it is evident that there will be certain underlying features especially crucial for distinguishing edges from non-edges. To further learn these subtle features from a small database, we adopt the concept in cost-sensitive learning. The original 2-way softmax training cost is the negative log-likelihood cost:

\begin{equation}
-\sum_i \log P(y_i \,|\, \bx_i, \btheta) =
-\sum_i y_i \log P(y_i=1 \,|\, \bx_i, \btheta)
-\sum_i (1-y_i) \log P(y_i=0 \,|\, \bx_i, \btheta)
\label{eqn:cost}
\end{equation}

\noindent where $\bx_i$ is the input image patch, $\btheta$ is the parameters of CNN, and $y_i$ is the binary ($0$ or $1$) edge label prediction. This cost is computed above the 2-way softmax layer, and will be back-propagated to train all convolutional layers. To apply cost-sensitive fine-tuning, we consider a biased negative log-likelihood cost:

\begin{equation}
-\sum_i \alpha\, y_i \log P(y_i=1 \,|\, \bx_i, \btheta)
-\sum_i \beta\, (1-y_i) \log P(y_i=0 \,|\, \bx_i, \btheta)
\label{eqn:biased}
\end{equation}

where $\alpha$ and $\beta$ are respectively the bias for positive (edge) or negative (non-edge) training data. If $\alpha=1$ and $\beta=1$, (\ref{eqn:biased}) is reduced to the original negative log-likelihood cost as in (\ref{eqn:cost}). In our approach, we set $\alpha=2\beta$ for positive cost-sensitive fine-tuning, and $2\alpha=\beta$ for negative cost-sensitive fine-tuning. Notice that, rather than directly back-propagating with (\ref{eqn:biased}), a convenient and alternative strategy is to create biased sampling for fine-tuning with (\ref{eqn:cost}). That is, for positive cost-sensitive fine-tuning, we sample twice more edge patches than non-edge ones, and vice versa, for negative cost-sensitive fine-tuning.

\begin{table}[tH]
\caption{(a) Contour detection results of employing different fine-tuning schemes. The experiments use only Conv5 features, since this setting reflects the effectiveness of fine-tuning directly. (b) Contour detection results of using CNN features from different layers. Conv1-3 denotes using features in convolutional layers 1 to 3, and Conv1-5, layers 1 to 5.  Note that, for the sake of comparisons, all models in (a) and (b) do not include features from the surrounding pixels as described in Section~\ref{ssec:baseline}.}
\label{tbl:joint}
\centering
\renewcommand{\arraystretch}{1.5}
\addtolength{\tabcolsep}{+8pt}
{\small
\begin{tabular}{@{}c@{\hskip 2.1cm}c@{}}
\raisebox{0.3cm}{
\begin{tabular}{@{}l@{\hskip 0.3cm}c@{\hskip 0.3cm}c@{\hskip 0.3cm}c@{}}
{\normalsize \textbf{(a)}} & ODS & OIS & AP \\ \Xhline{3\arrayrulewidth}
Pre-trained (baseline) & .604 & .620 & .546 \\
Traditional fine-tune & .573 & .585 & .524 \\
Per-pixel fine-tune & .620 & .632 & .561 \\
Positive fine-tune & .612 & .624 & .542 \\
Negative fine-tune & .624 & .639 & .566 \\
\textbf{Pos + Neg fine-tune} & .638 & .650 & .579 \\
\end{tabular}
}
&
\begin{tabular}{@{}l@{\hskip 0.3cm}c@{\hskip 0.3cm}c@{\hskip 0.3cm}c@{}}
{\normalsize \textbf{(b)}} & ODS & OIS & AP \\ \Xhline{3\arrayrulewidth}
Conv1 & .627 & .660 & .625 \\
Conv2 & .699 & .718 & .712 \\
Conv3 & .655 & .670 & .619 \\
Conv4 & .654 & .667 & .615 \\
Conv5 & .604 & .620 & .546 \\ \hline
Conv1-3 & .733 & .753 & .750 \\
Conv1-5 & .741 & .759 & .757 \\
\end{tabular}
\end{tabular}
}
\end{table}

\subsection{Final Fusion Model}
\label{ssec:final_model}

The overall framework is an ensemble model. We combine an ImageNet pre-trained model, a per-pixel fine-tuned model, a positive cost-sensitive fine-tuned model, and a negative cost-sensitive fine-tuned model together. We use a heuristic branch-and-bound scheme to decide the fusion coefficients. The idea of fusing different training models is to capture different aspects of features. It is worthy mentioning that the improvements owing to the model fusion indicates that the various fine-tunings have their own merits on feature learning and are all useful in this respect.

%
\section{Experiment Results}
\label{sec:exp}
%

We test our method on the Berkeley Segmentation Dataset and Benchmark (BSDS500) \citep{bsds,gpb}. To better assess the effects of the various fine-tuning techniques, we report their respective performance of contour detection. Comparisons with other competitive methods are also included to demonstrate the effectiveness of the proposed model.

The BSDS500 dataset is the current de facto standard image collection for contour detection. The dataset contains 200 training, 100 validation, and 200 testing images. Boundaries in each image are labeled by several workers and are averaged to form the ground truth. The accuracy of contour detection is evaluated by three measures: the best F-measure on the dataset for a fixed threshold (ODS), the aggregate F-measure on the dataset for the best threshold in each image (OIS), and the average precision (AP) on the full recall range \citep{gpb}. Prior to evaluation, we apply a standard non-maximal suppression technique to edge maps to obtain thinned edges \citep{canny}.

\subsection{On Fine-tuning}
\label{ssec:exp_finetune}

The parameter fine-tuning is done on a a server with a GeForce GTX Titan Black GPU card. We set the overall learning rate as tenth of the original ImageNet pre-trained learning rate, and the softmax learning rate as ten times of the overall learning rate. The modification to the proposed per-pixel fine-tuning speeds up the parameter fine-tuning process. It takes $3$ days to finish $100,000$ iterations of per-pixel fine-tuning, while requiring more than $10$ days for traditional fine-tuning. For both traditional fine-tuning and per-pixel fine-tuning, we sample $500$ boundary (edge) and $500$ non-boundary (non-edge) patches per training image. For positive cost-sensitive fine-tuning, we sample $1000$ boundary patches and $500$ non-boundary patches per training image, while $500$ boundary patches and $1000$ non-boundary patches per training image for negative cost-sensitive fine-tuning.

We report the results of the various fine-tuning techniques in Table~\ref{tbl:joint}(a). The experiments use only Conv5 features, and are carried out with SVM classifications. Since this setting is most similar to a softmax fine-tuning architecture, we can directly observe the effectiveness of fine-tuning. The experiment results show that, compared with the baseline (the pre-trained model), traditional fine-tuning, which is the original fine-tuning architecture with padding in every layer, degrades overall performance by $0.2$ to $0.4$. This implies that traditional per-image fine-tuning is not appropriate in learning per-pixel features for per-pixel applications. On the other hand, per-pixel fine-tuning improves the performance by about $0.15$ in all measurements. Pertaining to cost-sensitive fine-tuning, when compared with the per-pixel fine-tuning, positive fine-tuning slightly degrades and negative fine-tuning slightly improves. One possible explanation is that there are relatively more non-boundary regions than boundary points, so features learned for non-boundary regions improve the overall performance. However, if we combine features of positive and negative fine-tuning, the performance is significantly boosted again by $0.2$. The performance gain signifies the complementary property of positive and negative fine-tunings as expected.

In conclusion, per-pixel fine-tuning raises the performances of per-pixel applications. Also, the combination of positive and negative cost-sensitive fine-tunings improves the classification performance the most. Therefore, it supports the advantage of using an ensemble fine-tuning model.

\begin{table}[tH]
\caption{Comparisons with other methods on BSDS500. We compare our methods (5-stream Conv1-5 pre-trained model and final fusion model, denoted as CSCNN) with seven competitive techniques, including gPb-owt-ucm, Sketch Tokens, Sparse Code Gradients (SCG), DeepNet, Pointwise Mutual Information (PMI+sPb), $N^{4}$-fields and Structured Edges. Except DeepNet and $N^{4}$-fields, we also plot the PR curves of all compared methods.}
\label{tbl:comparison}
\centering
\renewcommand{\arraystretch}{1.5}
\addtolength{\tabcolsep}{+8pt}
{\small
\begin{tabular}{@{}c@{\hskip 0.8cm}c@{}}
\raisebox{2.8cm}{
\begin{tabular}{@{}l@{\hskip 0.3cm}c@{\hskip 0.3cm}c@{\hskip 0.3cm}c@{}}
Method & ODS & OIS & AP \\ \Xhline{3\arrayrulewidth}
gPb-owt-ucm  \citep{gpb}  & .73 & .76 & .73 \\
Sketch tokens  \citep{sketchtokens} & .73 & .75 & .78 \\
SCG  \citep{scg} & .74 & .76 & .77 \\
DeepNet  \citep{deepnet} & .74 & .76 & .76 \\
PMI+sPb, MS \citep{pmi} & .74 & .77 & .78 \\
$N^{4}$-fields \citep{n4} & .75 & .77 & .78 \\
SE-MS-HS \citep{se} & .75 & .77 & \textbf{.80} \\ \hline
5-stream pre-trained model & .75 & .77 & .77 \\
\textbf{Final fusion model} (CSCNN) & \textbf{.76} & \textbf{.78} & \textbf{.80}
\end{tabular}
}
&
\includegraphics[width=0.38\textwidth]{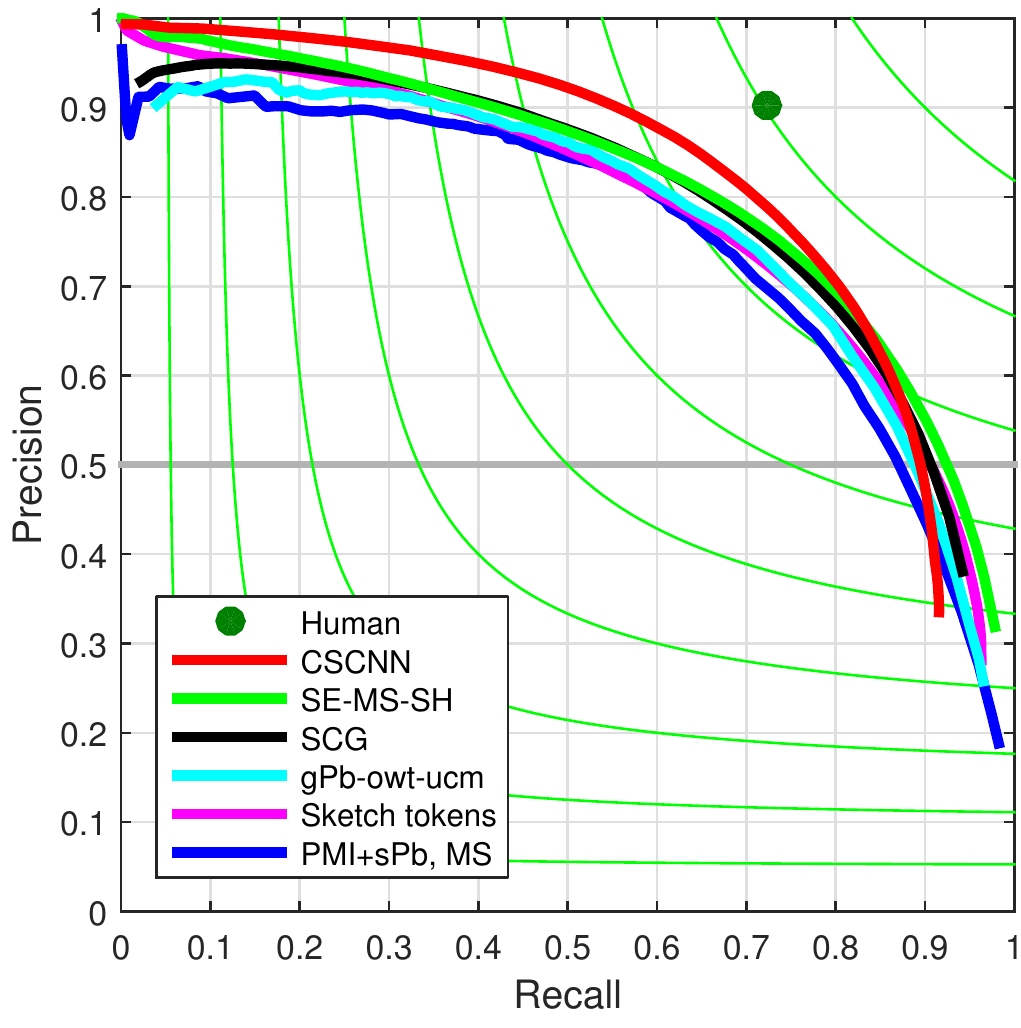}\\
\end{tabular}
}
\end{table}

\subsection{On Features in Different Layers}
\label{exp:layer}

We next conduct experiments to show how features from different convolutional layers contribute to the performance. In Table~\ref{tbl:joint}(b), we see that features in the second convolutional layer contribute the most, and then the third and the fourth layer. These suggest that low- to mid-level features are most useful for contour detection, while the lowest- and higher-level features provide additional boost. Although features in the first and the fifth convolutional layer are less effective when employed alone, we achieve the best results by combining all five streams. It indicates that the local edge information in low-level features and the object contour information in higher-level features are both necessary for achieving high performance in contour detection tasks.

\subsection{Contour Detection Results and Comparisons}
\label{exp:comp}

\begin{figure}
\begin{center}
\begin{tabular}{c@{\hskip 0.1cm}c@{\hskip 0.1cm}c@{\hskip 0.1cm}c@{\hskip 0.1cm}c}
\includegraphics[width=0.19\textwidth,frame]{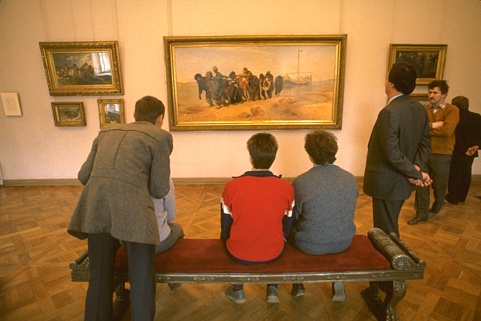} &
\includegraphics[width=0.19\textwidth,frame]{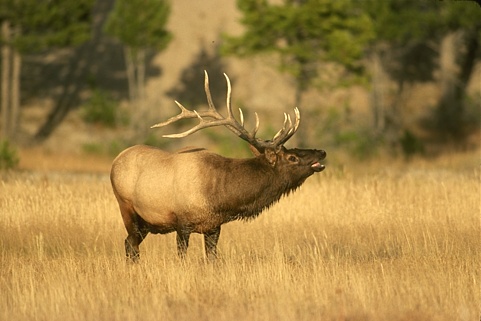}  &
\includegraphics[width=0.19\textwidth,frame]{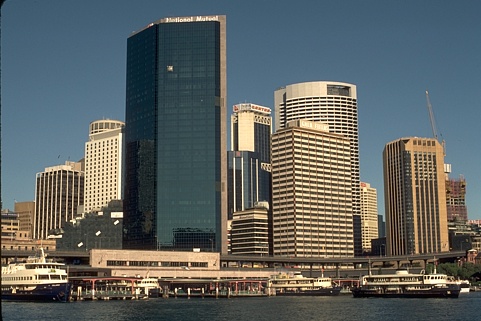}  &
\includegraphics[width=0.19\textwidth,frame]{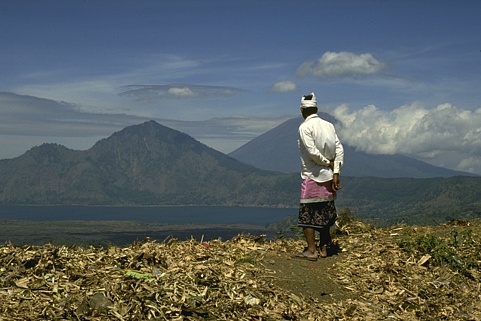} &
\includegraphics[width=0.19\textwidth,frame]{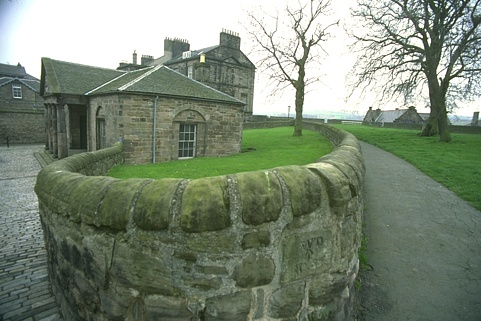} \\
\includegraphics[width=0.19\textwidth,frame]{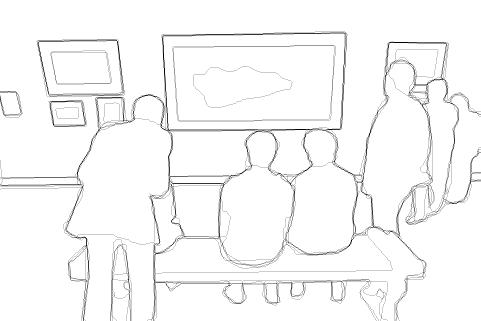} &
\includegraphics[width=0.19\textwidth,frame]{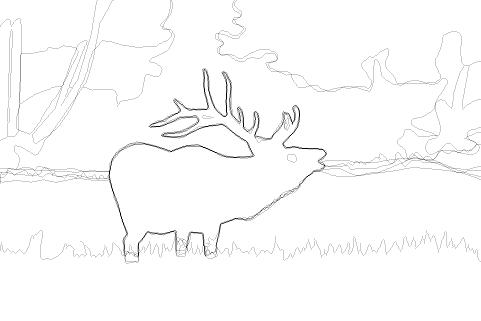}  &
\includegraphics[width=0.19\textwidth,frame]{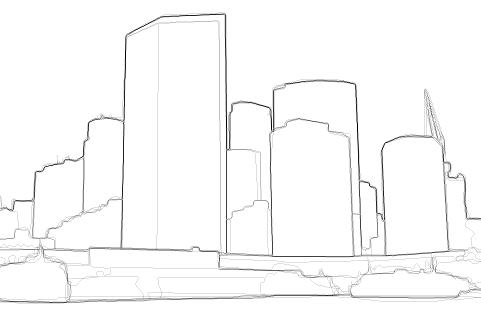}  &
\includegraphics[width=0.19\textwidth,frame]{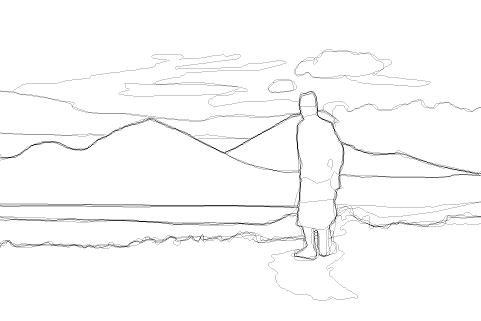} &
\includegraphics[width=0.19\textwidth,frame]{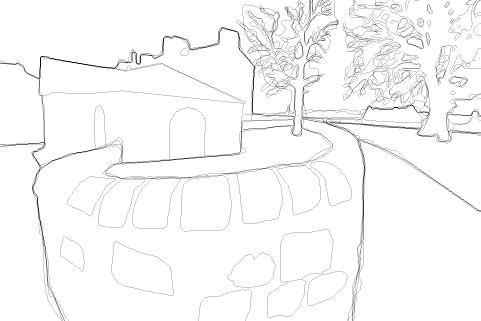} \\
\includegraphics[width=0.19\textwidth,frame]{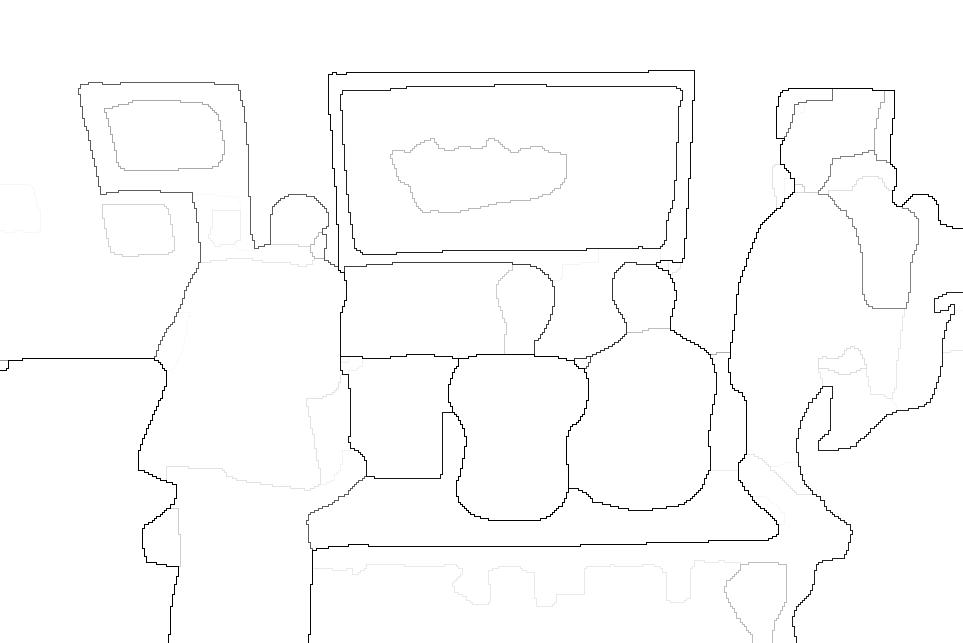} &
\includegraphics[width=0.19\textwidth,frame]{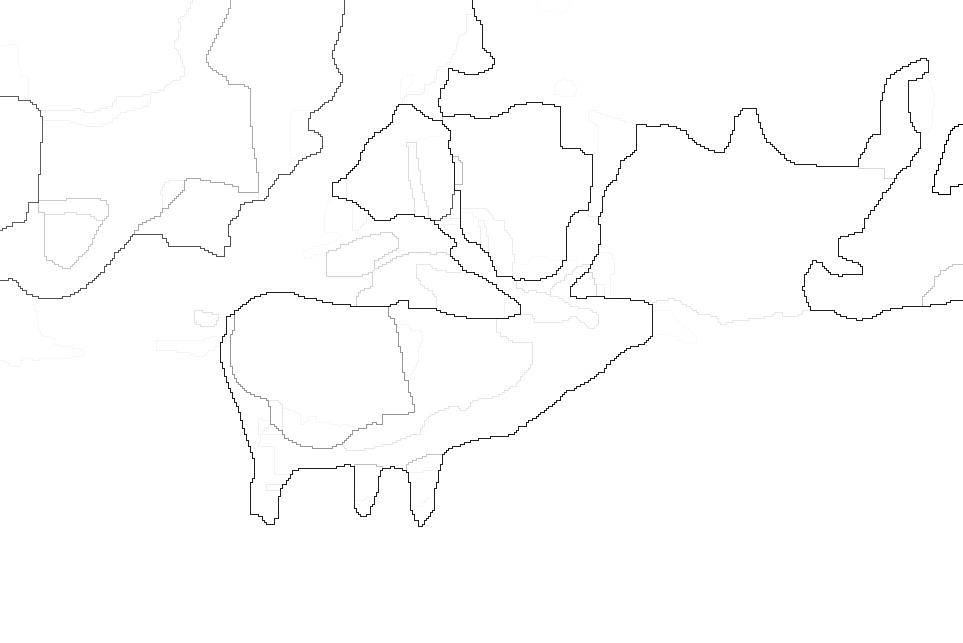}  &
\includegraphics[width=0.19\textwidth,frame]{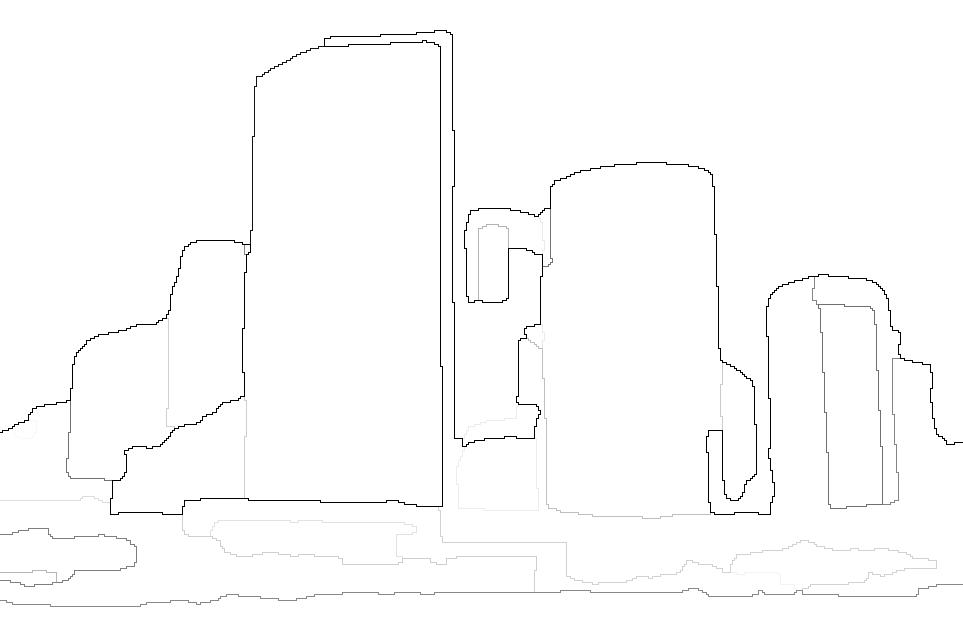}  &
\includegraphics[width=0.19\textwidth,frame]{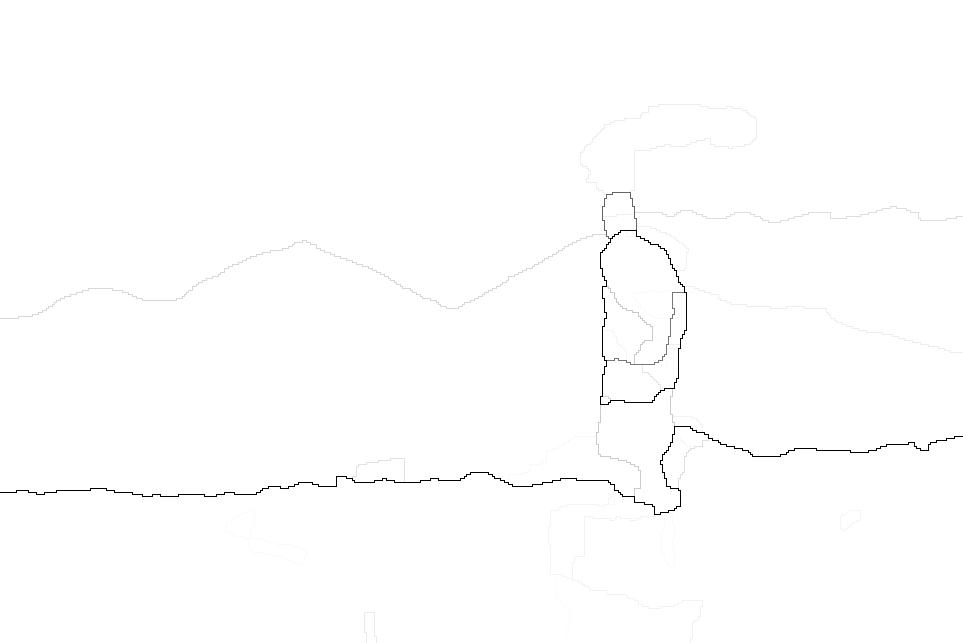} &
\includegraphics[width=0.19\textwidth,frame]{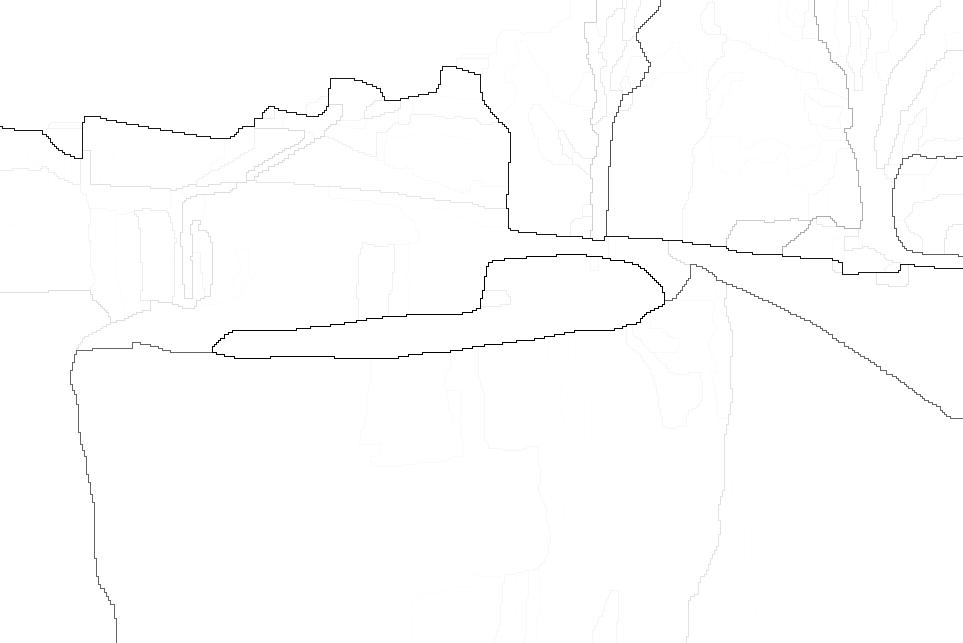} \\
\includegraphics[width=0.19\textwidth,frame]{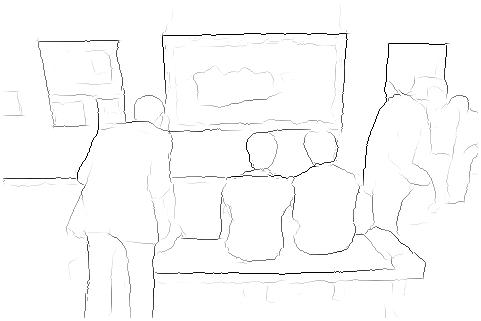} &
\includegraphics[width=0.19\textwidth,frame]{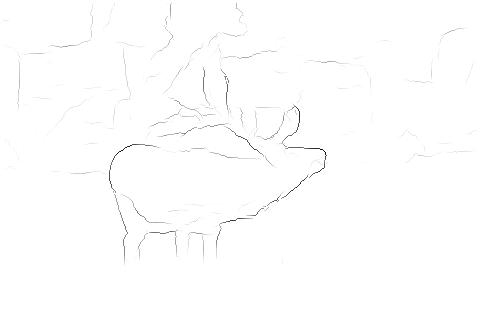}  &
\includegraphics[width=0.19\textwidth,frame]{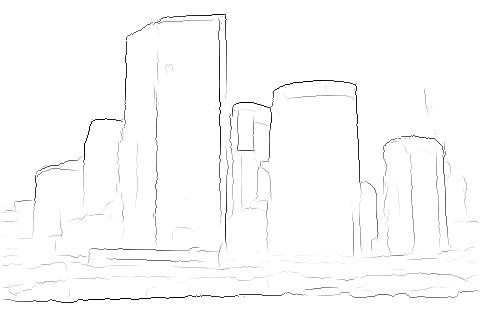}  &
\includegraphics[width=0.19\textwidth,frame]{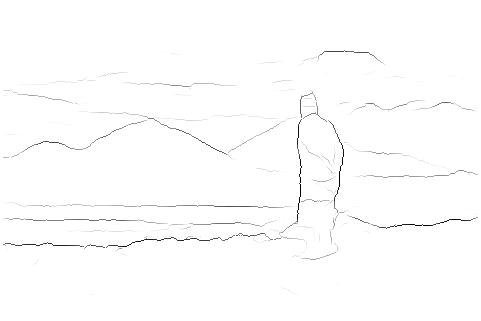} &
\includegraphics[width=0.19\textwidth,frame]{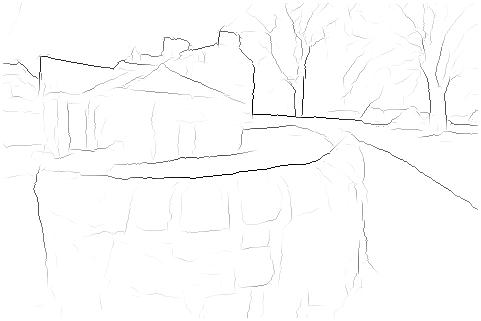} \\
\includegraphics[width=0.19\textwidth,frame]{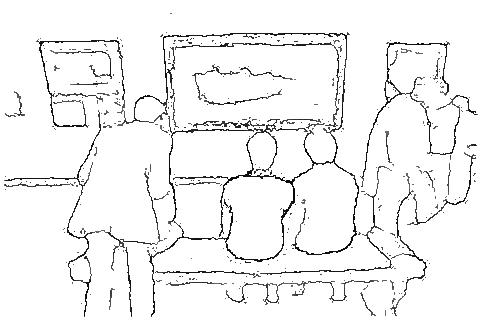} &
\includegraphics[width=0.19\textwidth,frame]{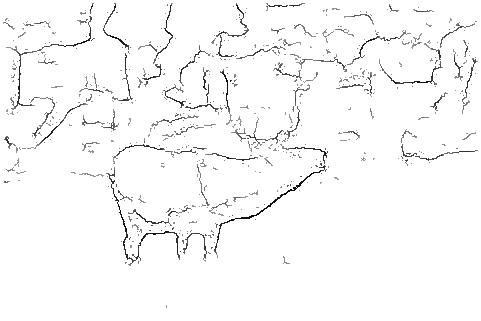}  &
\includegraphics[width=0.19\textwidth,frame]{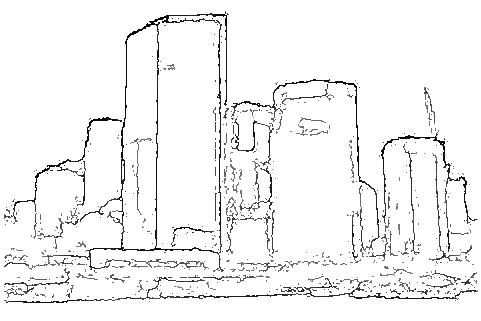}  &
\includegraphics[width=0.19\textwidth,frame]{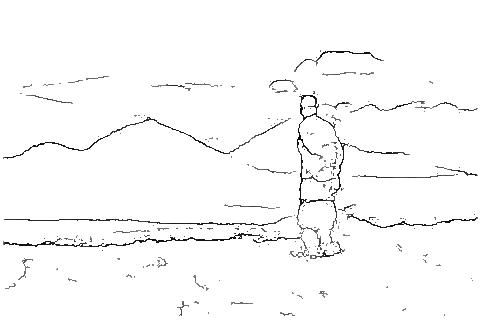} &
\includegraphics[width=0.19\textwidth,frame]{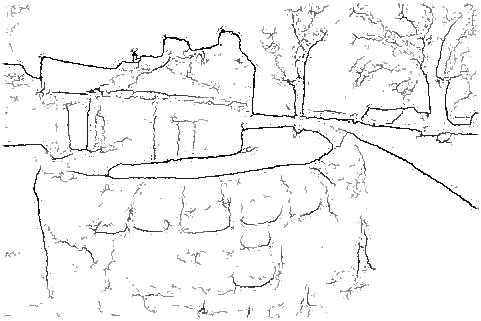} \\
\includegraphics[width=0.19\textwidth,frame]{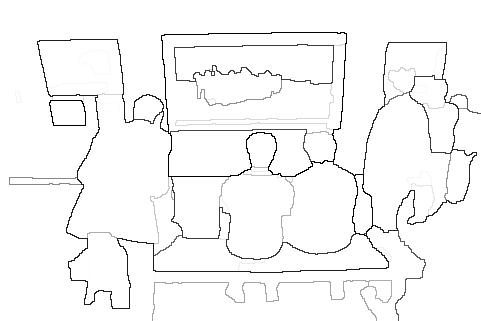} &
\includegraphics[width=0.19\textwidth,frame]{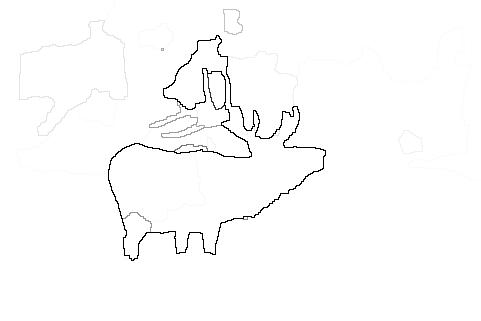}  &
\includegraphics[width=0.19\textwidth,frame]{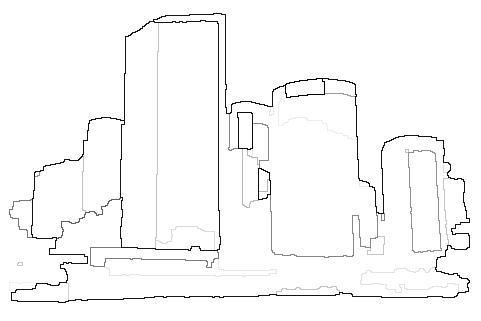}  &
\includegraphics[width=0.19\textwidth,frame]{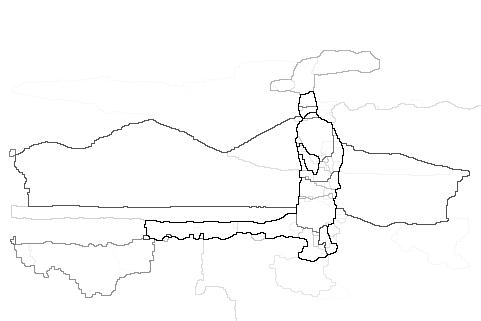} &
\includegraphics[width=0.19\textwidth,frame]{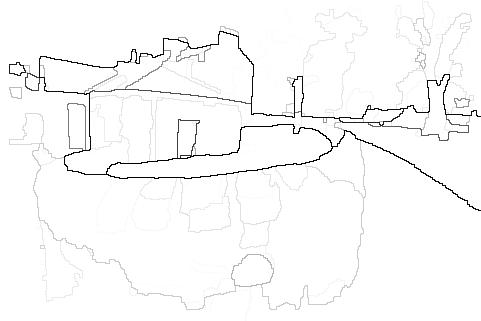} \\
\includegraphics[width=0.19\textwidth,frame]{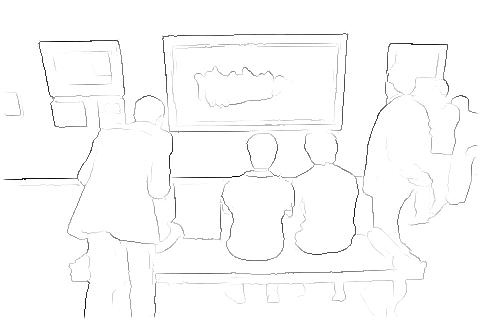} &
\includegraphics[width=0.19\textwidth,frame]{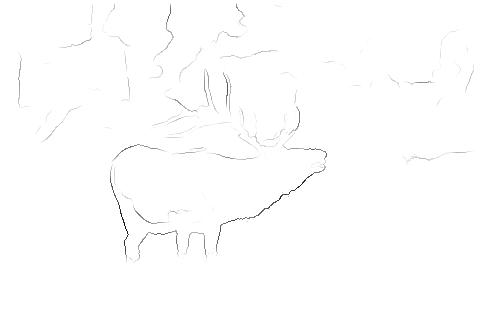}  &
\includegraphics[width=0.19\textwidth,frame]{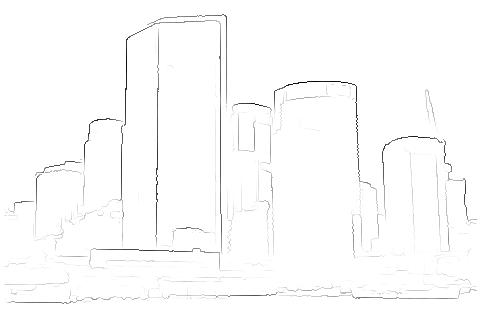}  &
\includegraphics[width=0.19\textwidth,frame]{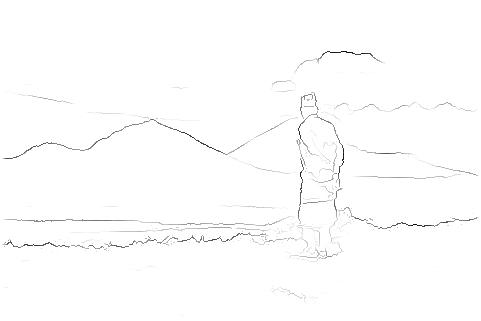} &
\includegraphics[width=0.19\textwidth,frame]{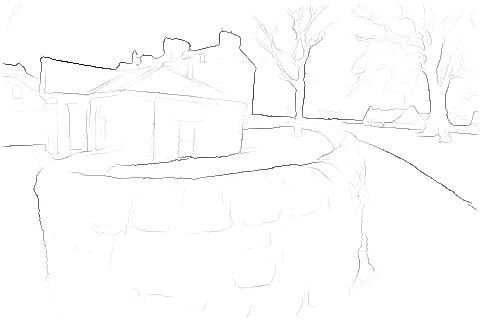} \\
\includegraphics[width=0.19\textwidth,frame]{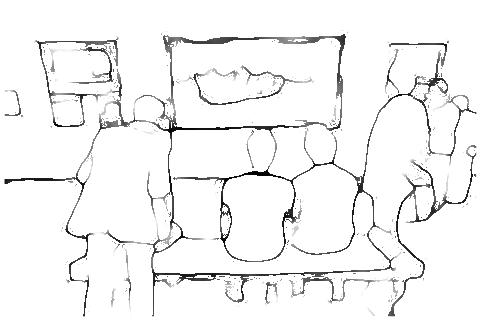} &
\includegraphics[width=0.19\textwidth,frame]{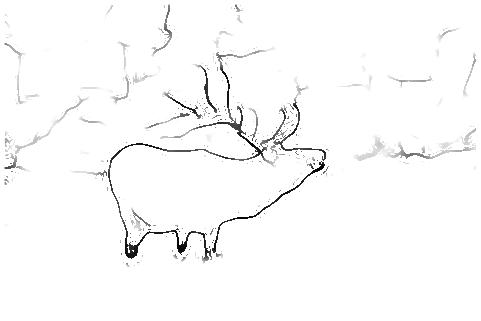}  &
\includegraphics[width=0.19\textwidth,frame]{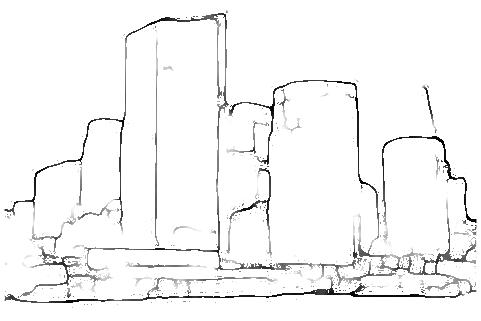}  &
\includegraphics[width=0.19\textwidth,frame]{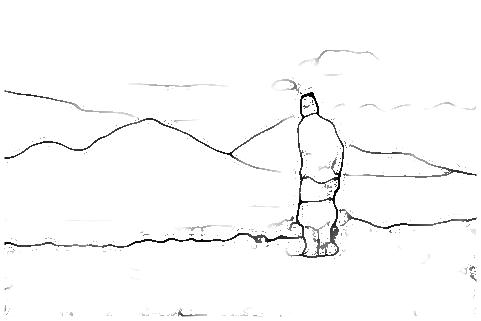} &
\includegraphics[width=0.19\textwidth,frame]{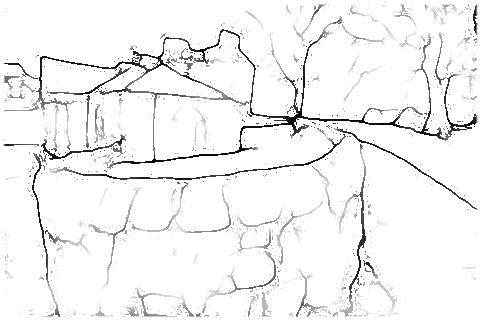} \\
\end{tabular}
\end{center}
\caption{Visualization results. From top to bottom: Input image, ground truth, gPb-owt-ucm, SCG, sketch tokens, PMI, SE-MS (MS: multiscale), CSCNN (our method, cost-sensitive CNN). We display edges above the best threshold for each image.}
\label{fig:visual}
\end{figure}

Finally, we show the experimental results of our pre-trained model and final fusion model. In Table~\ref{tbl:comparison}, we report the contour detection performances on BSDS500 by our methods and seven competitive techniques, including gPb \citep{gpb}, Sketch Tokens \citep{sketchtokens}, Sparse Code Gradients \citep{scg}, DeepNet \citep{deepnet}, Pointwise Mutual Information \citep{pmi}, $N^4$-fields \citep{n4} and Structured Edges \citep{se}. While our 5-stream ImageNet pre-trained model (using features from all five convolutional layers) already achieves impressive results for contour detection on ODS and OIS measurements, the proposed fine-tuning techniques can further improve the performance. In particular, the final ensemble model improves from $0.75$ to $0.76$ on ODS measurement, and from $0.77$ to $0.78$ on OIS measurement. It also achieves state-of-the-art performance on the AP measurement. In Figure~\ref{fig:visual}, we include a number of contour detection examples for qualitative visualization.

%
\section{Discussion}
\label{sec:discussion}
%

In this work, we describe how to use the DenseNet architecture to tailor for per-pixel computer vision problems, such as contour detection. We propose fine-tuning techniques to more effectively carry out parameter learning with a per-pixel based cost function and to overcome the limitation of using a small training set. The resulting cost-sensitive model appears to be promising for generating useful per-pixel feature vectors and should be useful for computer vision applications requiring analyzing local image property. An interesting future research direction is to establish a proper dimensionality reduction framework for the resulting high-dimensional per-pixel feature vectors and to examine its effects on the performance of contour detection.

\comment{
\subsubsection*{Acknowledgments}
Use unnumbered third level headings for the acknowledgments. All
acknowledgments, including those to funding agencies, go at the end of the paper.
}

\bibliography{CSCNN}

\begin{thebibliography}{34}
\providecommand{\natexlab}[1]{#1}
\providecommand{\url}[1]{\texttt{#1}}
\expandafter\ifx\csname urlstyle\endcsname\relax
  \providecommand{\doi}[1]{doi: #1}\else
  \providecommand{\doi}{doi: \begingroup \urlstyle{rm}\Url}\fi

\bibitem[Arbelaez et~al.(2011)Arbelaez, Maire, Fowlkes, and Malik]{gpb}
Arbelaez, Pablo, Maire, Michael, Fowlkes, Charless, and Malik, Jitendra.
\newblock Contour detection and hierarchical image segmentation.
\newblock \emph{Pattern Analysis and Machine Intelligence, IEEE Transactions
  on}, 33\penalty0 (5):\penalty0 898--916, 2011.

\bibitem[Bowyer et~al.(1999)Bowyer, Kranenburg, and Dougherty]{bowyer1999edge}
Bowyer, Kevin, Kranenburg, Christine, and Dougherty, Sean.
\newblock Edge detector evaluation using empirical roc curves.
\newblock In \emph{Computer Vision and Pattern Recognition, 1999. IEEE Computer
  Society Conference on.}, volume~1. IEEE, 1999.

\bibitem[Briggman et~al.(2009)Briggman, Denk, Seung, Helmstaedter, and
  Turaga]{briggman2009maximin}
Briggman, Kevin, Denk, Winfried, Seung, Sebastian, Helmstaedter, Moritz~N, and
  Turaga, Srinivas~C.
\newblock Maximin affinity learning of image segmentation.
\newblock In \emph{Advances in Neural Information Processing Systems}, pp.\
  1865--1873, 2009.

\bibitem[Canny(1986)]{canny}
Canny, John.
\newblock A computational approach to edge detection.
\newblock \emph{Pattern Analysis and Machine Intelligence, IEEE Transactions
  on}, \penalty0 (6):\penalty0 679--698, 1986.

\bibitem[Deng et~al.(2009)Deng, Dong, Socher, Li, Li, and
  Fei-Fei]{Deng2009imagenet}
Deng, Jia, Dong, Wei, Socher, Richard, Li, Li-Jia, Li, Kai, and Fei-Fei, Li.
\newblock Image{N}et: A large-scale hierarchical image database.
\newblock In \emph{CVPR}, pp.\  248--255, 2009.

\bibitem[Doll{\'a}r \& Zitnick(2014)Doll{\'a}r and Zitnick]{se}
Doll{\'a}r, Piotr and Zitnick, C~Lawrence.
\newblock Fast edge detection using structured forests.
\newblock \emph{arXiv preprint arXiv:1406.5549}, 2014.

\bibitem[Doll{\'a}r et~al.(2006)Doll{\'a}r, Tu, and
  Belongie]{dollar2006supervised}
Doll{\'a}r, Piotr, Tu, Zhuowen, and Belongie, Serge.
\newblock Supervised learning of edges and object boundaries.
\newblock In \emph{Computer Vision and Pattern Recognition, 2006 IEEE Computer
  Society Conference on}, volume~2, pp.\  1964--1971. IEEE, 2006.

\bibitem[Farabet et~al.(2013)Farabet, Couprie, Najman, and
  LeCun]{farabet2013learning}
Farabet, Clement, Couprie, Camille, Najman, Laurent, and LeCun, Yann.
\newblock Learning hierarchical features for scene labeling.
\newblock \emph{Pattern Analysis and Machine Intelligence, IEEE Transactions
  on}, 35\penalty0 (8):\penalty0 1915--1929, 2013.

\bibitem[Fram \& Deutsch(1975)Fram and Deutsch]{fram1975quantitative}
Fram, Jerry~R. and Deutsch, Edward~S.
\newblock On the quantitative evaluation of edge detection schemes and their
  comparison with human performance.
\newblock \emph{Computers, IEEE Transactions on}, 100\penalty0 (6):\penalty0
  616--628, 1975.

\bibitem[Ganin \& Lempitsky(2014)Ganin and Lempitsky]{n4}
Ganin, Yaroslav and Lempitsky, Victor.
\newblock {$N^{4}$}-fields: Neural network nearest neighbor fields for image
  transforms.
\newblock \emph{arXiv preprint arXiv:1406.6558}, 2014.

\bibitem[Girshick et~al.(2013)Girshick, Donahue, Darrell, and
  Malik]{girshick2013rich}
Girshick, Ross, Donahue, Jeff, Darrell, Trevor, and Malik, Jitendra.
\newblock Rich feature hierarchies for accurate object detection and semantic
  segmentation.
\newblock \emph{arXiv preprint arXiv:1311.2524}, 2013.

\bibitem[Hariharan et~al.(2014)Hariharan, Arbel{\'a}ez, Girshick, and
  Malik]{hariharan2014hypercolumns}
Hariharan, Bharath, Arbel{\'a}ez, Pablo, Girshick, Ross, and Malik, Jitendra.
\newblock Hypercolumns for object segmentation and fine-grained localization.
\newblock \emph{arXiv preprint arXiv:1411.5752}, 2014.

\bibitem[Iandola et~al.(2014)Iandola, Moskewicz, Karayev, Girshick, Darrell,
  and Keutzer]{densenet}
Iandola, Forrest, Moskewicz, Matt, Karayev, Sergey, Girshick, Ross, Darrell,
  Trevor, and Keutzer, Kurt.
\newblock Densenet: Implementing efficient convnet descriptor pyramids.
\newblock \emph{arXiv preprint arXiv:1404.1869}, 2014.

\bibitem[Isola et~al.(2014)Isola, Zoran, Krishnan, and Adelson]{pmi}
Isola, Phillip, Zoran, Daniel, Krishnan, Dilip, and Adelson, Edward~H.
\newblock Crisp boundary detection using pointwise mutual information.
\newblock In \emph{Computer Vision--ECCV 2014}, pp.\  799--814. Springer, 2014.

\bibitem[Jarrett et~al.(2009)Jarrett, Kavukcuoglu, Ranzato, and
  LeCun]{jarrett2009best}
Jarrett, Kevin, Kavukcuoglu, Koray, Ranzato, M, and LeCun, Yann.
\newblock What is the best multi-stage architecture for object recognition?
\newblock In \emph{Computer Vision, 2009 IEEE 12th International Conference
  on}, pp.\  2146--2153. IEEE, 2009.

\bibitem[Jia et~al.(2014)Jia, Shelhamer, Donahue, Karayev, Long, Girshick,
  Guadarrama, and Darrell]{caffe}
Jia, Yangqing, Shelhamer, Evan, Donahue, Jeff, Karayev, Sergey, Long, Jonathan,
  Girshick, Ross, Guadarrama, Sergio, and Darrell, Trevor.
\newblock Caffe: Convolutional architecture for fast feature embedding.
\newblock \emph{arXiv preprint arXiv:1408.5093}, 2014.

\bibitem[Kivinen et~al.(2014)Kivinen, Williams, Heess, and
  Technologies]{deepnet}
Kivinen, Jyri~J, Williams, Christopher~KI, Heess, Nicolas, and Technologies,
  DeepMind.
\newblock Visual boundary prediction: A deep neural prediction network and
  quality dissection.
\newblock In \emph{Proceedings of the Seventeenth International Conference on
  Artificial Intelligence and Statistics}, pp.\  512--521, 2014.

\bibitem[Krizhevsky et~al.(2012)Krizhevsky, Sutskever, and Hinton]{alexnet}
Krizhevsky, Alex, Sutskever, Ilya, and Hinton, Geoffrey~E.
\newblock Imagenet classification with deep convolutional neural networks.
\newblock In \emph{Advances in neural information processing systems}, pp.\
  1097--1105, 2012.

\bibitem[LeCun et~al.(1989)LeCun, Boser, Denker, Henderson, Howard, Hubbard,
  and Jackel]{lecun1989backpropagation}
LeCun, Yann, Boser, Bernhard, Denker, John~S, Henderson, Donnie, Howard,
  Richard~E, Hubbard, Wayne, and Jackel, Lawrence~D.
\newblock Backpropagation applied to handwritten zip code recognition.
\newblock \emph{Neural computation}, 1\penalty0 (4):\penalty0 541--551, 1989.

\bibitem[LeCun et~al.(1998)LeCun, Bottou, Bengio, and
  Haffner]{lecun1998gradient}
LeCun, Yann, Bottou, L{\'e}on, Bengio, Yoshua, and Haffner, Patrick.
\newblock Gradient-based learning applied to document recognition.
\newblock \emph{Proceedings of the IEEE}, 86\penalty0 (11):\penalty0
  2278--2324, 1998.

\bibitem[Lim et~al.(2013)Lim, Zitnick, and Doll{\'a}r]{sketchtokens}
Lim, Joseph~J, Zitnick, C~Lawrence, and Doll{\'a}r, Piotr.
\newblock Sketch tokens: A learned mid-level representation for contour and
  object detection.
\newblock In \emph{Computer Vision and Pattern Recognition (CVPR), 2013 IEEE
  Conference on}, pp.\  3158--3165. IEEE, 2013.

\bibitem[Long et~al.(2014)Long, Shelhamer, and Darrell]{long2014fully}
Long, Jonathan, Shelhamer, Evan, and Darrell, Trevor.
\newblock Fully convolutional networks for semantic segmentation.
\newblock \emph{arXiv preprint arXiv:1411.4038}, 2014.

\bibitem[Mairal et~al.(2008)Mairal, Leordeanu, Bach, Hebert, and
  Ponce]{mairal2008discriminative}
Mairal, Julien, Leordeanu, Marius, Bach, Francis, Hebert, Martial, and Ponce,
  Jean.
\newblock Discriminative sparse image models for class-specific edge detection
  and image interpretation.
\newblock In \emph{Computer Vision--ECCV 2008}, pp.\  43--56. Springer, 2008.

\bibitem[Malik et~al.(2001)Malik, Belongie, Leung, and Shi]{malik2001contour}
Malik, Jitendra, Belongie, Serge, Leung, Thomas, and Shi, Jianbo.
\newblock Contour and texture analysis for image segmentation.
\newblock \emph{International journal of computer vision}, 43\penalty0
  (1):\penalty0 7--27, 2001.

\bibitem[Martin et~al.(2001)Martin, Fowlkes, Tal, and Malik]{bsds}
Martin, David, Fowlkes, Charless, Tal, Doron, and Malik, Jitendra.
\newblock A database of human segmented natural images and its application to
  evaluating segmentation algorithms and measuring ecological statistics.
\newblock In \emph{Computer Vision, 2001. ICCV 2001. Proceedings. Eighth IEEE
  International Conference on}, volume~2, pp.\  416--423. IEEE, 2001.

\bibitem[Martin et~al.(2004)Martin, Fowlkes, and Malik]{martin2004learning}
Martin, David~R, Fowlkes, Charless~C, and Malik, Jitendra.
\newblock Learning to detect natural image boundaries using local brightness,
  color, and texture cues.
\newblock \emph{Pattern Analysis and Machine Intelligence, IEEE Transactions
  on}, 26\penalty0 (5):\penalty0 530--549, 2004.

\bibitem[Perona \& Malik(1990)Perona and Malik]{perona1990scale}
Perona, Pietro and Malik, Jitendra.
\newblock Scale-space and edge detection using anisotropic diffusion.
\newblock \emph{Pattern Analysis and Machine Intelligence, IEEE Transactions
  on}, 12\penalty0 (7):\penalty0 629--639, 1990.

\bibitem[Sermanet et~al.(2013{\natexlab{a}})Sermanet, Eigen, Zhang, Mathieu,
  Fergus, and LeCun]{sermanet2013overfeat}
Sermanet, Pierre, Eigen, David, Zhang, Xiang, Mathieu, Micha{\"e}l, Fergus,
  Rob, and LeCun, Yann.
\newblock Overfeat: Integrated recognition, localization and detection using
  convolutional networks.
\newblock \emph{arXiv preprint arXiv:1312.6229}, 2013{\natexlab{a}}.

\bibitem[Sermanet et~al.(2013{\natexlab{b}})Sermanet, Kavukcuoglu, Chintala,
  and LeCun]{sermanet2013pedestrian}
Sermanet, Pierre, Kavukcuoglu, Koray, Chintala, Soumith, and LeCun, Yann.
\newblock Pedestrian detection with unsupervised multi-stage feature learning.
\newblock In \emph{Computer Vision and Pattern Recognition (CVPR), 2013 IEEE
  Conference on}, pp.\  3626--3633. IEEE, 2013{\natexlab{b}}.

\bibitem[Shotton et~al.(2008)Shotton, Blake, and
  Cipolla]{shotton2008multiscale}
Shotton, Jamie, Blake, Andrew, and Cipolla, Roberto.
\newblock Multiscale categorical object recognition using contour fragments.
\newblock \emph{Pattern Analysis and Machine Intelligence, IEEE Transactions
  on}, 30\penalty0 (7):\penalty0 1270--1281, 2008.

\bibitem[Turaga et~al.(2010)Turaga, Murray, Jain, Roth, Helmstaedter, Briggman,
  Denk, and Seung]{turaga2010convolutional}
Turaga, Srinivas~C, Murray, Joseph~F, Jain, Viren, Roth, Fabian, Helmstaedter,
  Moritz, Briggman, Kevin, Denk, Winfried, and Seung, H~Sebastian.
\newblock Convolutional networks can learn to generate affinity graphs for
  image segmentation.
\newblock \emph{Neural Computation}, 22\penalty0 (2):\penalty0 511--538, 2010.

\bibitem[Xiaofeng \& Bo(2012)Xiaofeng and Bo]{scg}
Xiaofeng, Ren and Bo, Liefeng.
\newblock Discriminatively trained sparse code gradients for contour detection.
\newblock In \emph{Advances in neural information processing systems}, pp.\
  584--592, 2012.

\bibitem[Zheng et~al.(2010)Zheng, Yuille, and Tu]{zheng2010detecting}
Zheng, Songfeng, Yuille, Alan, and Tu, Zhuowen.
\newblock Detecting object boundaries using low-, mid-, and high-level
  information.
\newblock \emph{Computer Vision and Image Understanding}, 114\penalty0
  (10):\penalty0 1055--1067, 2010.

\bibitem[Zitnick \& Doll{\'a}r(2014)Zitnick and Doll{\'a}r]{zitnick2014edge}
Zitnick, C~Lawrence and Doll{\'a}r, Piotr.
\newblock Edge boxes: Locating object proposals from edges.
\newblock In \emph{Computer Vision--ECCV 2014}, pp.\  391--405. Springer, 2014.

\end{thebibliography}
\bibliographystyle{iclr2015}

\end{document}